\documentclass[pmlr,twocolumn,10pt]{jmlr} 

\usepackage{microtype}
\usepackage{graphicx}
\usepackage{booktabs} 
\usepackage{multirow}
\usepackage{cleveref}
\usepackage{hyperref}

\usepackage{subfiles}
\usepackage{siunitx}

\theorembodyfont{\upshape}
\theoremheaderfont{\scshape}
\theorempostheader{:}
\theoremsep{\newline}

\jmlrvolume{LEAVE UNSET}
\jmlryear{2023}
\jmlrsubmitted{LEAVE UNSET}
\jmlrpublished{LEAVE UNSET}
\jmlrworkshop{Conference on Health, Inference, and Learning (CHIL) 2023} 

\title[]{Explaining a machine learning decision to physicians via counterfactuals}

\author{%
\Name{Supriya Nagesh}\Email{nsupriy@amazon.com}\\
\addr Amazon 
\AND
\Name{Nina Mishra}\Email{nmishra@amazon.com}\\
\addr Amazon
\AND
\Name{Yonatan Naamad}\Email{ynaamad@amazon.com}\\
\addr Amazon
\AND
\Name{James M. Rehg}\Email{rehg@gatech.edu}\\
\addr Georgia Institute of Technology 
\AND
\Name{Mehul A. Shah}\Email{mehul@aryn.ai}\\
\addr Aryn
\AND
\Name{Alexei Wagner}\Email{awagner@bwh.harvard.edu}\\
\addr Harvard Medical School 
}

\begin{document}

\maketitle

\begin{abstract}

Machine learning models perform well on several healthcare tasks and can help reduce the burden on the healthcare system. However, the lack of explainability is a major roadblock to their adoption in hospitals.  
\textit{How can the decision of an ML model be explained to a physician?} 
The explanations considered in this paper are counterfactuals (CFs), hypothetical scenarios that would have resulted in the opposite outcome. Specifically, time-series CFs are investigated, inspired by the way physicians converse and reason out decisions `I would have given the patient a vasopressor if their blood pressure was lower and falling'. 
Key properties of CFs that are particularly meaningful in clinical settings are outlined: physiological plausibility, relevance to the task and sparse perturbations. 
Past work on CF generation does not satisfy these properties, specifically plausibility in that realistic time-series CFs are not generated.
A variational autoencoder (VAE)-based approach is proposed that captures these desired properties.
The method produces CFs that improve on prior approaches quantitatively (more plausible CFs as evaluated by their likelihood w.r.t original data distribution, and 100$\times$ faster at generating CFs)
and qualitatively (2$\times$ more plausible and relevant) 
as evaluated by three physicians.

\end{abstract}

\paragraph*{Data and Code Availability}
This paper uses the MIMIC-III dataset~\citep{johnson2016mimic}
which is available on the Physionet repository~\citep{goldberger2000physiobank}. Our implementation of CF VAE is available at \href{https://github.com/supriyanagesh94/CFVAE}{https://github.com/supriyanagesh94/CFVAE}.

\paragraph*{Institutional Review Board (IRB)}
Our research does not require IRB approval.

\section{Introduction}
\label{sec:intro}

Machine learning (ML) models are demonstrating compelling performance on various healthcare tasks ~\citep{futoma2017learning,kaji2019attention,johnson2017reproducibility}, 
and even reach physician-level performance in some cases~\citep{rajpurkar2017chexnet,grewal2018radnet,hannun2019cardiologist}. Machine learning can particularly 
be useful in reducing the burden on the healthcare system by predicting hospital bed usage~\citep{kutafina2019recursive,turgeman2017insights},
intervention use~\citep{suresh2017clinical,ghassemi2017predicting}, and rehospitalization~\citep{henry2015targeted}. 
Despite this, we have not seen widespread adoption of these models in hospitals, with explainability being a key reason~\citep{amann2020explainability}. Physicians want to understand how a model produced its output (a diagnosis or a recommendation for the course of treatment) before using it in practice.

In this paper, we explain a model's output through counterfactuals (CFs)~\citep{wachter2017counterfactual} - \textit{what should have changed in the input, to have a different outcome under this model?}. Our approach was inspired by observing our medical collaborators express clinical judgements in terms of hypothetical trends in vital signs, e.g., ``This patient would have been moved out of the ICU had their blood pressure been stable." 
The CFs provide a means for a physician to interrogate and understand the assessments of an ML model.

\paragraph{Desirable Properties of CFs}
What makes one CF better than another?
Three key properties desirable of CFs for use in clinical settings include:
\begin{enumerate}
    \item Plausibility: Generated CFs must be consistent with the patient population and not contain data that is  unlikely or impossible (e.g., diastolic blood pressure exceeding systolic).  
    \item Relevance: The CFs should reflect the key dimensions that are related to the task that the ML model is targeting. For example, given a model for the task of predicting if a patient will need a ventilator - a generated CF that is relevant might differ in the SpO2 and respiration rate trajectories (and not for example in their bilirubin level).    
    \item Sparse perturbations: The generated CFs must deviate minimally (minimal number of features) from the original patient to result in the opposite prediction. 
\end{enumerate}
The remainder of this paper presents a method for generating CFs possessing these properties.

\begin{figure}[h]
    \centering
    \includegraphics[scale=0.45]{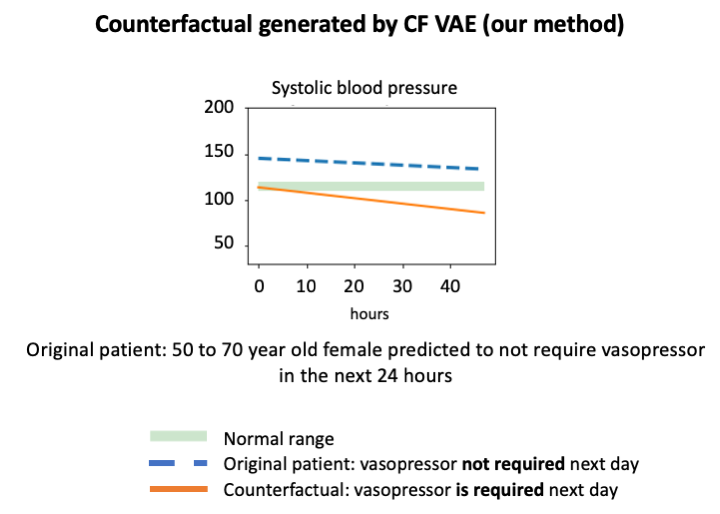}
    \caption{An intervention prediction model predicts that this patient will not need a vasopressor. The orange time series represents the CF generated in response to a physician interrogating the prediction. The dotted blue line is the actual time series and the green box shows the normal range.}
    \label{fig:intro}
\end{figure}

To illustrate a CF on real data generated by our approach, see Figure~\ref{fig:intro}. Given a patient's 48 hour vital sign history, an ML model predicts if they will require a vasopressor in the next 24 hours. We use our method to generate a CF given the learned model.
While the original data consists of multiple features (see Fig.~\ref{fig:cfviz}), only one is changed in the counterfactual. Hence, the counterfactual is sparsely perturbed in that it did not needlessly alter other time series.  Furthermore, the counterfactual is plausible in the sense that it is clinically possible for systolic blood pressure to drop from 100 to 70 over the course of 48 hours.  Finally, the counterfactual is relevant since vasopressors are administered when blood pressure is low.

While there are existing methods for generating numerical CFs~\citep{mothilal2020explaining,joshi2019towards,delaney2020instance}, we show that they can produce unrealistic CFs,
making them unsuitable for generating explanations in a healthcare setting. 
For the same patient in Figure~\ref{fig:intro}, DICE~\citep{mothilal2020explaining} generates a CF (Figure~\ref{fig:intro-dice}) with SpO2 declining from 98 to -98, and heart rate dropping from 60 to -60. In reality, negative values are not possible. Further, neither clinical metric is relevant since vasopressors are administered when blood pressure drops.
Note that while~\citep{delaney2020instance} demonstrates results on generating CF ECG time series, their methods can produce unrealistic counterfactuals in other health settings. We further discuss this in Sec.~\ref{sec:discussion}.

\begin{figure*}[h]
    \centering
    \includegraphics[scale=0.16]{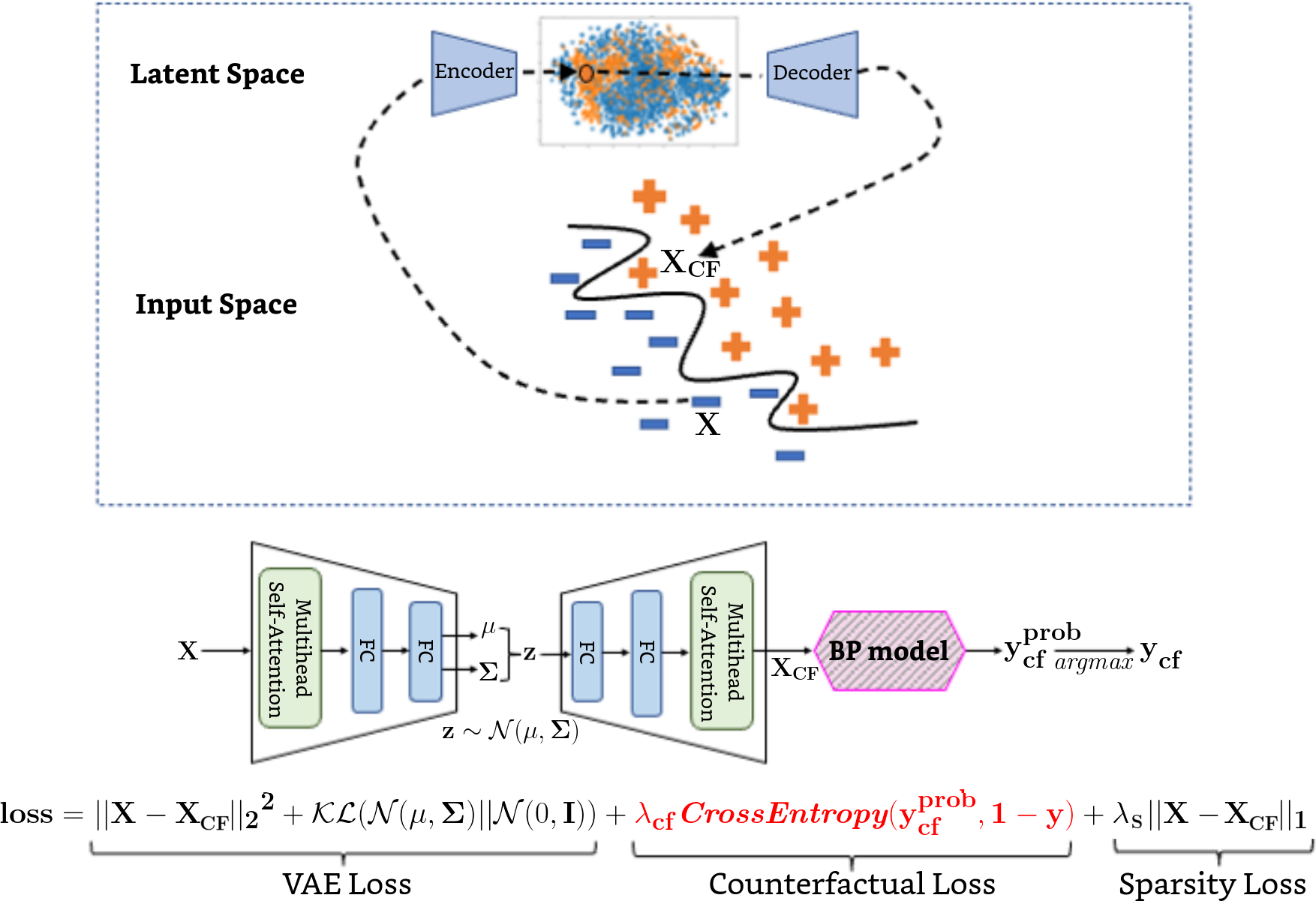}
    \caption{Our proposed approach: Given a binary prediction (BP) model, we learn a VAE latent space that embeds input data points onto regions of the latent space, which results in reconstructions that have the opposite label under the BP model.}
    \label{fig:intro_method}
\end{figure*}

In this paper, we develop CF-VAE - a method to produce counterfactuals based on variational autoencoders (VAE)~\citep{kingma2013auto}. Traditionally, VAEs are used to learn a continuous latent space and generate realistic synthetic data. We show how the traditional VAE loss function can be modified to generate convincing CFs. The CF must be close to the original patient's data (Fig.~\ref{fig:intro_method} first term), be of the opposite class (second term) and yet not perturb too many features (third term). As with VAEs, we are able to use stochastic gradient descent to train our model to generate desirable CFs. 

Additionally, CF-VAE is a feed-forward approach  - generating CFs only requires a feed-forward pass. In contrast, prior methods perform \textit{optimization at test time} to identify CF samples, as exemplified by ~\citep{mothilal2020explaining},~\citep{delaney2020instance},~\citep{delaney2020instance}, and ~\citep{joshi2019towards}
where generating CFs required making multiple calls to an optimization module.


In summary, this paper makes the following contributions: 1. We develop CF-VAE: a variational autoencoder based feed-forward method to produce plausible, relevant, and sparsely perturbed CFs, 2. We train an intervention prediction model for two ICU interventions: vasopressor and ventilator using the MIMIC-III clinical dataset. We generate CFs for each of these prediction models using CF-VAE, 3. We present a quantitative evaluation of the CFs produced by our method and prior works and note that a higher fraction of our CFs are plausible. 4. We present results from a qualitative analysis of CFs performed by three physicians. The physicians were blinded to the method that produced the CF and evaluated if a CF was plausible and relevant to the intervention. 
Our evaluators found that CF-VAE was 2X more plausible and relevant than DICE.


\section{Related Work}

Explainable machine learning is a heavily studied field too vast to adequately summarize in this section. Comprehensive surveys on the topic include ~\citep{doshi2017towards,adadi2018peeking,abdul2018trends,tjoa2020survey, chaddad2023survey}. 

Several explainability techniques focus on feature weights/attributions~\citep{zhang2019should,lundberg2017unified}. For example, suppose we have a learned model predicting if a person will be granted a loan. Analyzing the feature weights might reveal that the model attributes a larger weight to the feature `credit history' than the feature `marital status'. Feature attribution methods are useful in many settings.   

This paper differs in that the objective is to identify a CF to explain the model's decision~\citep{wachter2017counterfactual,joshi2019towards,mothilal2020explaining,delaney2020instance,xu2022counterfactual}.

Our approach to generating CFs over numerical data possesses three key properties not previously found in combination among prior numerical CF approaches: (1) Plausibility, (2) Relevance to the ML task, (3) Sparse perturbations. 
We compare the prior work on generating counterfactuals to explain a blackbox model~\citep{dhurandhar2018explanations,joshi2019towards,delaney2020instance,mothilal2020explaining,xu2022counterfactual}
 on each desirable characteristic we listed above. 

\textbf{{Plausibility:}} Our CFs are usually \textit{physiologically plausible}, meaning that we avoid reporting SpO$_2$ values of 105\% or diastolic blood pressures exceeding systolic since we sample from the latent space of a VAE trained on real patients. This contrasts with methods that look for nearby CFs under the Euclidean metric~\citep{ates2021counterfactual,guidotti2020explaining,dhurandhar2018explanations, mothilal2020explaining}, which are prone to sampling from outside the true data manifold. E.g., while an SpO$_2$ reading of 105 is close to a patient with SpO$_2$ 98, the former is not realistic. \citep{delaney2020instance} partially avoid this issue by using a nearest neighbor method. After identifying the nearest unlike neighbor (NUN) to a given time series,~\citep{delaney2020instance} changes the `high importance' regions of the input with segments from the NUN. While often sensible, this CF generation approach has a major pitfall when it comes to plausibility: in many cases, one cannot directly substitute a time series segment from one patient with that of another patient. \citep{joshi2019towards} ensures plausibility by training a VAE to approximate the data distribution and learn an embedding (latent) space. The algorithm optimizes a loss function to search over the latent space for a counterfactual.
\citep{xu2022counterfactual} posits that their method of only changing subsets which are both sufficient and necessary inherently induces plausibility. 

\textbf{{Relevance:}} Our CF-VAE incorporates a loss which captures relevance to the prediction task (that the ML model is trained on), encouraging CFs that are relevant to the prediction task. Whereas none of the other works explicitly encourage capturing representations that are relevant to the prediction task. 

\textbf{{Sparse perturbations:}} Our approach produces \textit{sparse perturbations} which can be interpreted as the factors important for the model outcome. We achieve sparse perturbations through a $\ell_1$ regularization on the perturbation. The work of~\citep{delaney2020instance} defines sparsity in terms of the length of the modified data segment rather than in terms of number of features they touch. Thus, they attempt to find short perturbations to the time series that flip the label. While this fits nicely within their framework, it is specific to signals whose label is largely defined by short-duration events (e.g. a single arrhythmia) and is not as well suited for sparsely capturing long-term trends in a small number of signals. 
In~\citep{mothilal2020explaining}, the authors propose augmenting their base model with an explicit post-hoc sparsification step. To sparsify the generated counterfactuals, they perform a greedy coordinate descent until no further coordinates can be independently minimized without crossing the decision boundary. Note that this post-hoc process can result in implausible CFs.
Because~\citep{xu2022counterfactual} works on a binary input domain, sparsity and relevance coincide in their work and so their search for a small set of diagnosis codes that collectively flips the classifier output inherently induces sparsity. The remaining works did not specifically target sparsity. 

Additionally, our CF-VAE can produce counterfactuals for time series data. Among the previously mentioned approaches,~\citep{delaney2020instance} is the only work explicitly generating time series counterfactuals. CF VAE produces CFs by performing a feed-forward pass through the encoder and decoder unlike~\citep{delaney2020instance,mothilal2020explaining,dhurandhar2018explanations,joshi2019towards} which require optimization at test-time to generate a CF.

\begin{table*}[h]
\small
    \centering
    \begin{tabular}{|l|c|c|c|c|c|c|}
          \hline
           & CEM &  REVISE &  DICE & NG-CF  &  CF-VAE (ours) \\
           \hline
            Plausibility  & {$\text{\sffamily X}$} &  {$\checkmark$} &  {$\text{\sffamily X}$} & {$\text{\sffamily X}$}  & {$\checkmark$} \\
         \hline
           Relevance  & {$\text{\sffamily X}$} &  {$\text{\sffamily X}$} &  {$\text{\sffamily X}$} & {$\text{\sffamily X}$}  & {$\checkmark$} \\
         \hline
            {{Sparse perturbations}} & {$\checkmark$}  & {$\text{\sffamily X}$} & {$\checkmark$}   & {$\text{\sffamily X}$}  & {$\checkmark$} \\
         \hline
           Time series  & {$\text{\sffamily X}$} &  {$\text{\sffamily X}$} &  {$\text{\sffamily X}$} & {$\checkmark$}  & {$\checkmark$} \\
         \hline
        Feed-forward approach &  {$\text{\sffamily X}$} & {$\text{\sffamily X}$} & {$\text{\sffamily X}$} & {$\text{\sffamily X}$}  & {$\checkmark$} \\ 
        \hline
    \end{tabular}
    \caption{\small{Comparison of different CF generation methods: CEM \citep{dhurandhar2018explanations}, REVISE \citep{joshi2019towards}, DICE \citep{mothilal2020explaining}, NG-CF \citep{delaney2020instance}, CF-VAE - this paper.}}
    \label{tab:salesman_matrix}
\end{table*}

\section{CF-VAE: Counterfactual Variational Autoencoder}
\label{sec:cfvae}

Given a target patient's data, our goal is to generate a CF patient for visualization as a means to explain how the model produced its prediction. Our solution comprises of a novel  Counterfactual VAE (CF-VAE) module (Fig.~\ref{fig:intro_method}), which provides a general, feed-forward approach to synthesizing CFs for time series classification problems.  

\begin{figure}[h]
    \centering
    \includegraphics[scale=0.45]{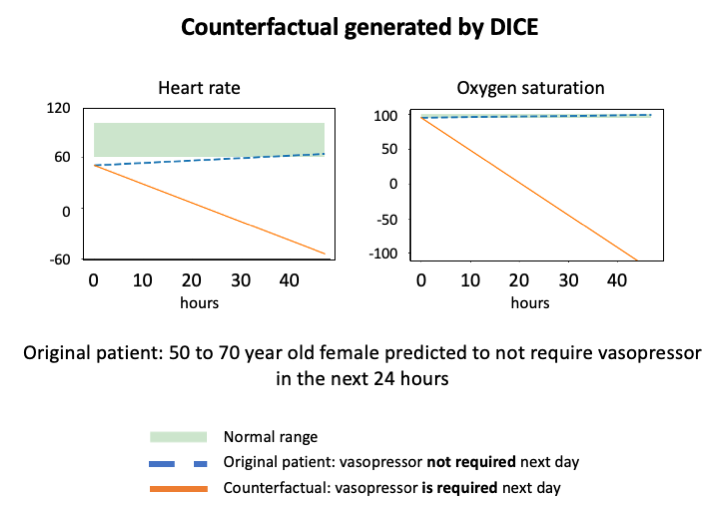}
    \caption{DICE's counterfactual for the same patient in Figure~\ref{fig:intro}}
    \label{fig:intro-dice}
\end{figure}

Our method (CF-VAE) jointly optimizes for plausibility (samples respect the data distribution), CF validity (samples flip the outcome of the ML model), and sparsity (minimal feature change). We show experimentally in Sec.~\ref{sec:results} that our joint training method produces more plausible CFs than prior methods~\citep{mothilal2020explaining,joshi2019towards,delaney2020instance}. A benefit of our method is that the embeddings learned by CF-VAE are \textit{relevant} to the prediction task, facilitating the generation of relevant CFs. It is difficult to achieve this type of relevance in the case of prior methods that do not utilize a task-specific representation during CF generation. An example of an irrelevant CF produced by DICE is shown in Figure~\ref{fig:intro-dice}. By incorporating multihead self-attention blocks~\citep{vaswani2017attention} in the encoder and decoder, the VAE can handle time series measurements as inputs, and synthesize counterfactual time series as outputs, thereby achieving the goals in Table~\ref{tab:salesman_matrix}. We now describe our solution architecture, beginning with a brief overview of a vanilla VAE \citep{kingma2013auto}.

\paragraph{VAE background:} The VAE approximation takes the form of a standard encoder-decoder pair where the encoder, $Q$, and the decoder, $P$, are each parameterized by neural networks. The encoder and decoder networks are trained by \textit{maximizing} the objective:
\begin{equation}
    \mathbb{E}_{\mathbf{X} \sim D}[\mathbb{E}_{\mathbf{z} \sim Q} [  \log{\mathcal{P}(\mathbf{X}|\mathbf{z})}] - \mathcal{D}(Q(\mathbf{z}|\mathbf{X}) || \mathcal{P}(\mathbf{z}))  ]
    \label{eq:vae}
\end{equation}

Where $\mathbf{X}$ is a data point sampled from the dataset $D$, the encoder $Q$ produces a distribution $\mathcal{N}(\mathbf{\mu_X},\mathbf{\Sigma_X})$ over the latent representation $\mathbf{z}$, and $\mathcal{D}$ is the KL divergence between the latent multivariate Gaussian distribution and the prior distribution $\mathcal{P}(\mathbf{z})$. The two terms in the objective function correspond to the reconstruction error and latent space normalization, respectively.  

When both the prior $\mathcal{P}(\mathbf{z})$ and output distributions $\mathcal{P}(\mathbf{X}|\mathbf{z})$ are assumed to be spherical Gaussians, maximizing \ref{eq:vae} can be shown to be equivalent to minimizing 
\begin{equation}
    \mathbb{E}_{\mathbf{X} \sim D}[{||\mathbf{X} - \mathbf{X'}||_2}^2 + \mathcal{KL}(\mathcal{N}(\mathbf{\mu_X}, \mathbf{\Sigma_X})|\mathcal{N}(\mathbf{0},\mathbf{I}))],
    \vspace{0.03in}
    \label{eq:vae2}
\end{equation}
where $\mathbf{X'} = P(Q(\mathbf{X}))$ is the network's reconstruction of $\mathbf{X}$.  We therefore use \ref{eq:vae2} as a starting point for defining a loss function for training our VAE. See \citep{doersch2016tutorial} for a more complete derivation of this objective. 

\newcommand{\BP}{\textrm{BP}}

\paragraph{CF-VAE objective:} To produce realistic CFs, we must generate samples with high probability under the data distribution that flip the output of a target binary prediction model (e.g., ICU intervention prediction).

We denote the binary prediction output of the target model as $y=\BP(\mathbf{X})$ and the output of the CF-VAE as $\mathbf{X_{CF}}$.

We modify the VAE objective so that its output ($\mathbf{X_{CF}}$) is penalized by a term proportional to the cross entropy of $y^{prob}_{cf}$ and $1-y$ (where $y^{prob}_{cf}$ is the class probability output of $\BP(\mathbf{X_{CF})}$) in addition to the standard regularization and reconstruction loss on $\mathbf{X}$ and $\mathbf{X_{CF}}$. Introducing this extra loss term allows the VAE to learn about the target model's decision boundary and incentivizes it to synthesize a CF sample $\mathbf{X_{CF}}$ whose output $y_{cf}$ is of the opposite class.  
Intuitively, as diagrammed in Fig.~\ref{fig:intro}(b), this teaches the VAE to encode the classifier boundary in its latent space, and to map a given $\mathbf{X}$ to a latent point of the opposite class. Another consideration is to have \textit{minimal} changes to the input to produce a CF. We achieve this through a sparsity constraint on the perturbation. Thus, our modified loss function takes on the form
\begin{align}
    \mathbb{E}_{\mathbf{X} \sim D}[ & {||\mathbf{X} - \mathbf{X_{CF}}||_2}^2 +  \mathcal{KL}(\mathcal{N}(\mathbf{\mu_X}, \mathbf{\Sigma_X})|\mathcal{N}(\mathbf{0},\mathbf{I})) \nonumber \\
   &  + \lambda_{cf} \textrm{CrossEntropy}(y^{prob}_{cf}, 1-y) \nonumber \\
   & + \lambda_{S}||\mathbf{X} - \mathbf{X_{CF}}||_1 ] 
\label{eq:cfvae}
\end{align}

where $\mathbf{X_{CF}}$ is the decoder output and $\lambda_S$, $\lambda_{cf}$ 
are loss weights
for sparsity and for the ``soft constraint'' that CF class $y_{cf}$ and $y$ must differ, respectively. Note that Eq.~\ref{eq:cfvae} differs from the standard VAE loss only in the cross-entropy and $\ell_1$ norm term. Adjusting $\lambda_{cf}$ allows us to tune the VAEs attention between focusing on its reconstruction/regularization objectives and on its CF objective. Fig.~\ref{fig:tsne} visualizes the latent space using t-SNE with varying $\lambda_{cf}$ -- as we increase $\lambda_{cf}$, we see more separation between the classes in the latent space. We choose the $\lambda_S$ value proportional to the magnitude of the different loss terms on the training set. See Sec.~\ref{sec:app_sparsity} for examples of CFs generated with and without the sparsity term. We provide intuition about the loss function by building it up one term at a time in Appendix~\ref{sec:app_loss_interpretation}.

A strength of our approach is that the CF-VAE can be trained in the same way as a Vanilla VAE, using stochastic gradient descent, allowing us to leverage the VAE optimization literature. Note that the parameters of the binary prediction model are held fixed while training the CF-VAE.   

We explore two ways to represent patient data: 1) A temporal representation based on linear trends (e.g., up or down) with a specific slope and intercept over the measurement window, 2) As a time series with a $N \times T$ data matrix, where $T$ is the number of time samples and $N$ is the number of measurements (e.g. vital signs). 
We present results for both patient representations in the subsequent sections.

\section{Experiments and Results}
\label{sec:results}

We use data from the MIMIC III ICU dataset for our experiments. Additional details are provided in Appendix~\ref{sec:app_data}. 
The main objectives of our experiments are to show that CF-VAE produces plausible, relevant, and sparsely perturbed CFs as explanations.

\subsection{Binary prediction task}
\label{sec:bp_task}

In our experiments, we generate CFs for two target binary prediction models - predicting the use of vasopressor and ventilator. Given a patient's 48 hour history, we train a model to predict if a target intervention will be required within the next 24 hours.  See Appendix~\ref{sec:app_modeldetails} for model implementation details. 
The results on the task of intervention prediction using the linear (slope-intercept) representation is shown in Table~\ref{tab:rank_results}. We perform additional experiments using the entire temporal sequence data as input to the prediction model. These results are presented in Table~\ref{tab:full_rank_results} of Appendix~\ref{sec:app_hh}.

The specific task of intervention prediction was motivated by the clinical problem of home hospital~\citep{levine2021hospital} and we provide more context on this problem in the Appendix~\ref{sec:app_hh}.

\subsection{Baseline methods for generating counterfactuals}
\label{sec:baselines}

The space of prior CF methods can be roughly partitioned into three approaches: (1) Optimization-based approaches in the input space~\citep{dhurandhar2018explanations,mothilal2020explaining}, (2) Optimization-based approaches in the latent space of a generative model~\citep{joshi2019towards}, and (3) Input perturbation-based approaches~\citep{delaney2020instance}. The third category is not suitable for us since substituting a part of a patient's vital signs with vital signs from another patient could be unrealistic. We discuss this further in Sec.~\ref{sec:disc_delaney}.
In our experiments, we compare to DICE~\citep{mothilal2020explaining} and REVISE~\citep{joshi2019towards} to represent the two optimization categories.  

\subsection{Counterfactuals generated by CF-VAE}
\label{sec:counterfactuals}
\begin{figure*}[h]
    \centering
    \includegraphics[scale=0.5]{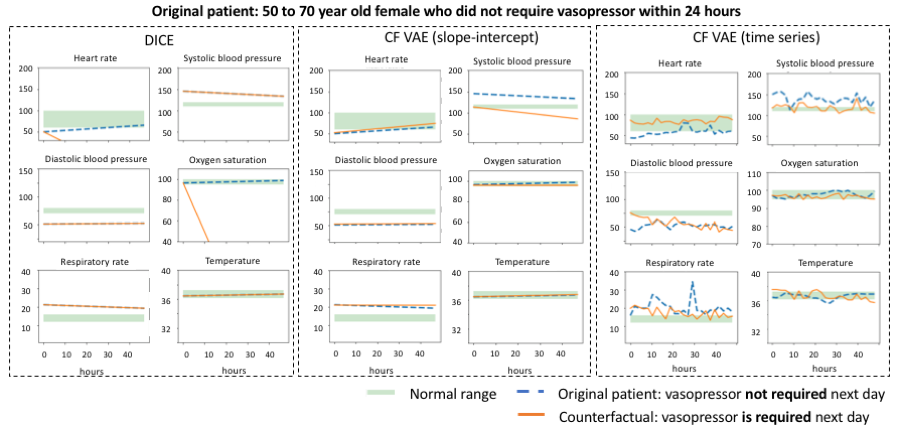}
    \caption{{Counterfactual (CF) generated using DICE \citep{mothilal2020explaining} and CF-VAE. The original patient (blue dotted) does not require a vasopressor within 24 hours. The orange line shows the generated CF. (a) CF generated using DICE: neither looks plausible nor relevant. Vasopressors are provided when the blood pressure drops - the DICE CF changes the heart rate and oxygen saturation. Note that the sharp drop in oxygen saturation is not realistic. The DICE CF is neither realistic nor relevant. (b) CF-VAE output with slope-intercept representation (c) CF-VAE output with a time series representation. In both (b), (c) CF-VAE is capturing the relationship between systolic BP, heart rate and vasopressor. }}
    \label{fig:cfviz}
\end{figure*}

We use the intervention prediction model as the binary prediction model and train CF-VAE to generate CFs given a test patient. The CFs produced for the  task of vasopressor prediction by CF-VAE and DICE and shown in \Cref{fig:cfviz}. We see that the CF-VAE produces a CF that changes the systolic blood pressure - which is a key trigger for providing vasopressors. The CF indicates that the binary prediction model has learned patterns between decreasing systolic blood pressure and the need for vasopressors. 
We quantitatively evaluate the CFs generated by CF-VAE to those generated by DICE and REVISE on three aspects:

\begin{table*}[h]
    \centering
    \begin{small}
    \begin{tabular}{l | c c c | c c c | c c }
    \toprule
     & \multicolumn{3}{c}{Ventilator} & \multicolumn{3}{c}{Vasopressor} & &  \\
    \hline
    
         \small{Method} & \small{\% $l_{\textrm{method}}$}  & \small{\% CF}  & \small{Physician}&  \small{\% $l_{\textrm{method}}$}  & \small{\% CF} & \small{Physician} & \small{Time (s)} & \small{Time (s)} \\
         &  \small{$\geq l_{\textrm{CF-VAE}}$} & \small{validity} & \small{score} &  \small{$\geq l_{\textrm{CF-VAE}}$} & \small{validity} & \small{score} & \small{(train)} & \small{(test)} \\
         \hline
         DICE & 2\% & 100\% & 40\%   & 2\% & 100\%  & 23.33\% & 0 & 0.37 \\
         REVISE & 16\% & 25\% & -   & 12\% & 19\% & -  & 10 & 0.38 \\
         CF-VAE & - & 90\% & 60\%   & - & 85\% & 73.33\%  & 180 & 0.001 \\
        \bottomrule
    \end{tabular}
    \end{small}
    \caption{{Comparison of \citep{mothilal2020explaining,joshi2019towards} and CF-VAE. $l_{\textrm{method}} > l_{\textrm{CF-VAE}}$ is the fraction of test points where the log likelihood of a CF from prior work exceeded that of our own. We see that in the case of Ventilator, only 2\% of the test CF from DICE had a higher likelihood score than our CF. The physician evaluated our CF for plausibility and relevance. The physician's score is the fraction of CF that were deemed both \textit{plausible} and \textit{relevant}. 
    Note that we don't have physician score for REVISE because of its poor validity percentage.}}
    \label{tab:kde_comparison}

\end{table*}

\paragraph{1. Log likelihood score under a KDE model:} We compute the log likelihood score of a generated CF under the kernel density estimator (with Gaussian kernel) fit to the training data to quantify its plausibility. A higher log-likelihood score implies that the CF is plausible and similar to a real patient in the training data. The column $\% l_{\textrm{method}} > l_{\textrm{CF-VAE}}$ in Table~\ref{tab:kde_comparison} is the ratio of test samples for which the likelihood score of CFs generated by DICE and REVISE were greater than that of CFs generated by CF-VAE, we outperform both these methods.

\paragraph{2. Validity of CFs generated:} Out of the total set of test points, Table~\ref{tab:kde_comparison} presents the percentage of generated points that have the opposite outcome under the binary prediction model (i.e. the true CF). Note that REVISE has a very low \% validity because the optimization did not converge to a CF.   

\paragraph{3. Train and test time:} The training and test time for each method is listed in the table. The train time is a one-time cost, while the test time listed is the average time for generating a CF for one test point. Since DICE and REVISE are optimization-based approaches, they incur a higher cost at test time. However, the cost of CF-VAE is a one-time training cost. Generating a CF with CF-VAE involves performing a forward pass though the trained model, which is 100x faster than DICE and REVISE. 

In addition to these measures, we evaluate the proximity of CFs generated at test time. Proximity ensures that we do not produce a very different synthetic patient as a CF for a given target patient. We observe that CF-VAE generates CFs that are closer to the input compared to DICE (Appendix~\ref{sec:app_proximity}).

\subsection{Qualitative evaluation of counterfactuals}
\label{sec:expert}

We presented 10 CFs generated by DICE and CF-VAE to three physicians to score based on plausibility and relevance. The physicians were blinded to whether the CF came from DICE vs. our method. A \textit{plausible and relevant} CF is one which convinces the physician that ``if the patient looked as in the CF, their intervention prediction would be reversed". 
The physicians evaluated CFs for two interventions - ventilator and vasopressor. Each physician evaluated 40 CFs, resulting in a total of 120 evaluated CFs.
On average, the three physicians marked 73.33\% and 60\% of the counterfactuals produced by CF-VAE and 23.33\% and 40\% of the counterfactuals produced by DICE for the vasopressor and ventilator prediction tasks respectively (Table~\ref{tab:kde_comparison}).
We exclude REVISE from this evaluation due to its poor convergence rate (it fails to produce a CF 75\% of the time). 

Compared to the two baselines, our method generates highly valid CFs that are more plausible and relevant (based on the small sample of expert evaluation), and does so in less time.

{\begin{table}[h]
    \centering
    \begin{tabular}{c c c }
    \toprule
     {} & Vaso & Vent \\
        \hline 
   ACC & $0.83\pm0.01$ & $0.83\pm0.12$ \\
   AUC & $0.88\pm0.02$ & $0.91\pm0.01$ \\
 
\bottomrule
\end{tabular}
\caption{{\small{Intervention prediction results using the slope-intercept representation for vasopressor and ventilator interventions. \textit{Margin of error generated by running the experiment with 10 random seeds.} Here, ACC = Accuracy, AUC = Area under ROC curve.} Additional results of prediction under three scenarios 1. Using temporal input representation, 2. Feature ablation on history of intervention, and 3. Including only patients who have received acute interventions at least once are presented in the appendix.
}}
\label{tab:rank_results}
\end{table}}

\section{Discussion}\label{sec:discussion}

\paragraph{Comparison to NG CF \citep{delaney2020instance}:}
\label{sec:disc_delaney}

\begin{figure}[h]
    \centering
    \includegraphics[scale=0.25]{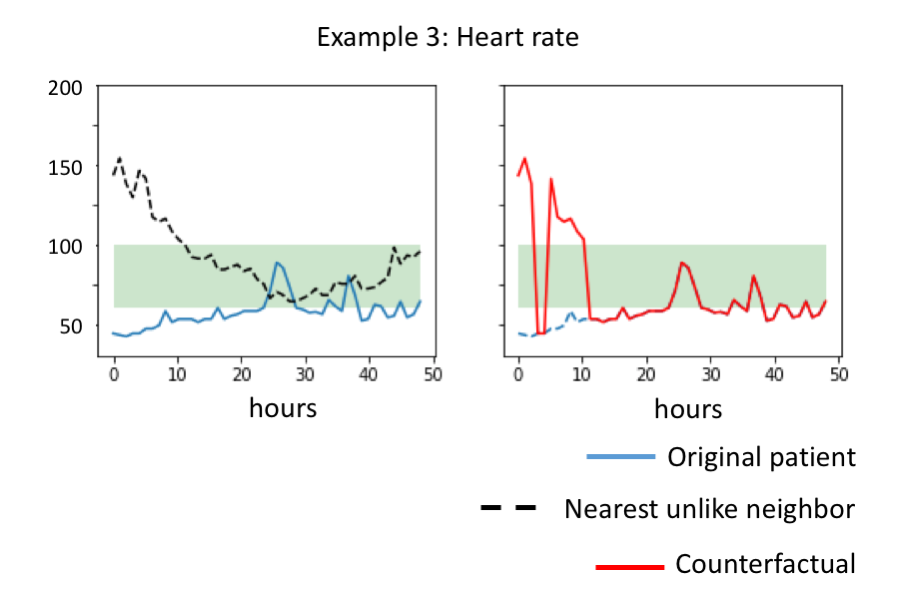}
    \caption{Illustration of the method in \citep{delaney2020instance}. The nearest unlike neighbor (NUN) to the target patient is identified. Segments of the time series of the NUN is stitched into the time series of the target patient. }
    \label{fig:delaney3}
\end{figure}

While~\citep{delaney2020instance} presents a method to produce counterfactual ECG signals, their method is not suitable for a general healthcare context. We briefly describe the method to generate a CF using the method of NG CF~\citep{delaney2020instance} here. Given a patient whose time series (e.g., heart rate) is shown in blue (Fig.~\ref{fig:delaney3}), the nearest unlike neighbor (NUN) in the training set is identified (shown in black here). Parts of the original time series are replaced with the NUN values resulting in a counterfactual as shown in red. Notice how the heart rate drops precipitously from 150 to 50 beats per minute. 
The NG CF method was proposed in the context of ECG signals. We believe that this could work for ECG signals (which have been studied in \citep{delaney2020instance}) - replacing a segment (e.g, the QRS complex) of the ECG signal of one patient with the QRS complex of another patient. However, given the amount of variability that exists in vital signs data, \textit{the strategy of stitching together time series from different patients could result in physiologically implausible counterfactuals.} Fig.~\ref{fig:delaney} is an example of a CF generated using NG CF. In comparison, CF-VAE produces plausible time series CF. See Fig.~\ref{fig:cfviz} for an example of a time series CF generated using CF-VAE. 

We compare the validity and run time of \citep{delaney2020instance} with our time series CF-VAE method. We see that \citep{delaney2020instance} has a 100\% validity since the method is designed to at least return the nearest unlike neighbor in the training set as a counterfactual. The training time of CF-VAE with time series is higher compared to \citep{delaney2020instance}, which does not require training. However, CF-VAE is ~80$\times$ faster than \citep{delaney2020instance} while generating a CF at test time.

\paragraph{Impact of sparsity term on the CFs:}
\label{sec:app_sparsity}

\begin{figure*}[h]
    \centering
    \includegraphics[scale=0.4]{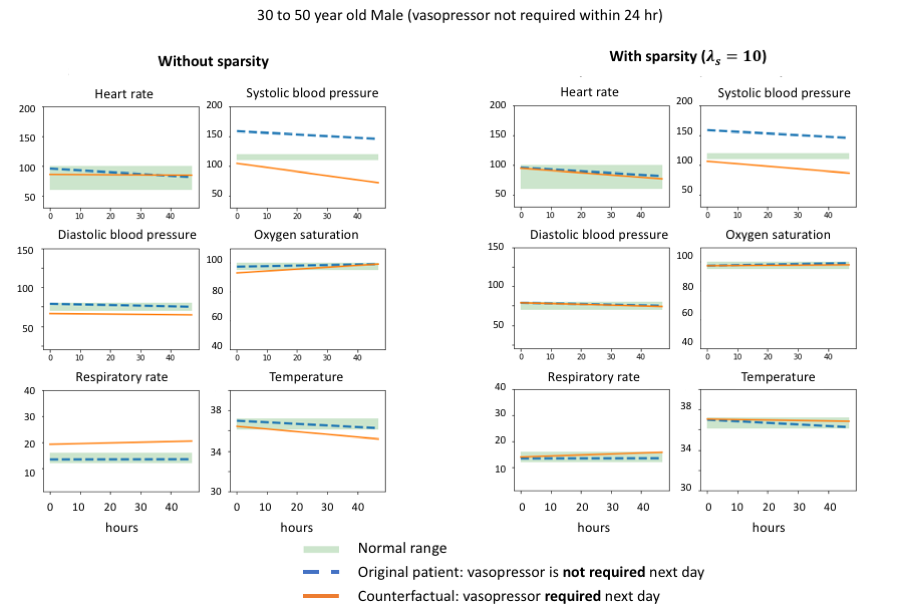}
    \caption{Example 1: Counterfactual produced with and without the sparsity term. Notice how only the systolic blood pressure changes when we include the sparsity constraint. Another example is shown in Fig.~\ref{fig:sparsity2}}
    \label{fig:sparsity1}
\end{figure*}

By sparsity in perturbation, we mean that a minimal number of features should be changed.  This is an important consideration as it reduces the cognitive load on a physician. We include a soft constraint to encourage CFs that alter a minimal number of dimensions with the $\lambda_{s} ||X - X_{CF}||_1 $ term in Eq.~\ref{eq:cfvae}. Figures~\ref{fig:sparsity1} and~\ref{fig:sparsity2} illustrate two examples with and without sparsity in the CF-VAE loss function. In both of these examples, we see that without sparsity, the CF-VAE produces a CF that changes multiple input dimensions - systolic blood pressure, temperature and respiratory rate. However, adding the sparsity constraint results in a CF that makes large changes only in the systolic blood pressure. Note that a vasopressor is administered to increase a patient's blood pressure up to normal levels. The CF-VAE with sparsity is correctly identifying the key feature relevant to the target intervention.

\begin{figure}[h!]
    \centering
    \includegraphics[scale=0.25]{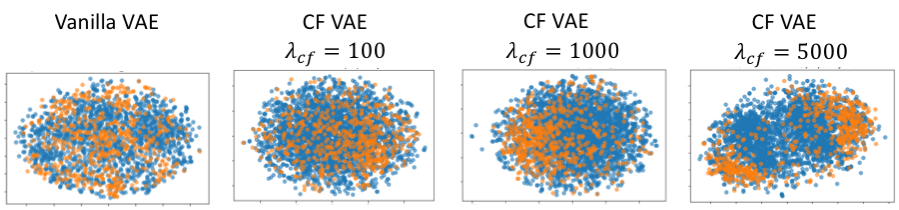}
    \caption{TSNE plot of the latent space of a vanilla VAE and CF-VAE with varying $\lambda_{cf}$. We visualize the two classes each point belongs to: requires intervention, or doesn't require intervention. We see that the CF-VAE captures the classifier boundary and learns to separate the two classes in the latent space as $\lambda_{cf}$ increases.}
    \label{fig:tsne}
\end{figure}

\paragraph{Visualization of the latent space of CF-VAE:}
We present the TSNE visualization of the latent space with varying values of $\lambda_{cf}$ in Figure \ref{fig:tsne}. The two classes (requires intervention, doesn't require intervention) are shown in two different colors. Notice that as the value of $\lambda_{cf}$ increases, the separation between the two classes becomes more pronounced - more information from the binary prediction model gets embedded in the latent space. 

\section{Significance}

Our motivation for pursuing the counterfactual problem came from `home hospital' programs, where patients are sent home to receive care they otherwise would have received in the hospital \citep{levine2021hospital}. The fluctuating availability of hospital beds and fears of hospital-acquired infection inspires interest in home hospital programs both within and outside of the US~\citep{knight2021provision}. 
Home hospital is an interesting solution where all parties benefit.  Patients are motivated by home hospital care: attractive benefits of such a program include improved sleep, home-cooked food, and avoiding hospital-acquired infections.  Hospitals are motivated to send less-acute patients home to expand bed capacity for more-acute patients. Past studies show that home hospital care tends to be substantially less expensive than in-hospital -- one study suggests 52\% cheaper~\citep{levine2021hospital}.  Moreover, early studies~\citep{levine2021hospital, leff2005hospital} suggest that home hospital care enjoys similar to slightly better outcomes, although these studies are limited in the number and variety of participants.  Findings confirm that home patients tend to be more physically active, sleep better, and have fewer bed sores. Note that there is global interest (not just US interest) in home hospital programs~\citep{knight2021provision} due to the sometimes diminished availability of hospital beds and fear of infection.

Selecting the right patients to be sent for home care is vital for the success of this program, e.g., patients who might require {\em acute} interventions (those that can only be performed in a hospital) must not be sent home. The current process of assigning a patient to home hospital relies on manual workflows of physicians constantly reviewing data, which is laborious and not scalable. Machine learning algorithms can be used to learn effective representations from large datasets containing patient records to identify and rank candidates based on their suitability for home hospital care. \textit{However, they need to be {explainable} in order to gain a physician's trust and be deployed in hospitals}. Hence, in this paper, we present a method to explain the decision of a trained ML model through counterfactuals generated by CF VAE. Given a model that is trained to determine if a patient can be sent to home hospital, our CF VAE can be used to generate counterfactuals and explain the decision of the model.

\section{Future work}

\citep{mothilal2020explaining} argue that diversity of CF is an important characteristic of a CF generation method. Diverse CFs alter \textit{different} feature dimensions to reverse the classifier outcome. A potential weakness of our method is that sampling from a smooth latent space may reduce diversity. There is a trade-off between diversity and ease of interpretation and this is an interesting topic for future work.

Causality is an interesting direction for future work.
The method in this paper is designed to minimally change an input point to move across a classification boundary. However, the factors that change are not necessarily causal.

Physicians already look at so many clinical time series, our CFs add even more. While our sparsity constraint is designed to reduce the number of time series, we may still be increasing the cognitive burden. One way to mitigate is with a simple English sentence, e.g., ``If this patient's blood oxygen had been 80\% and falling, and their respiration rate had been 20 bpm and rising, we would have predicted that a ventilator would be required in the next 12 hours.'' Translating these numeric time series CFs to meaningful, accurate English sentences is a good future direction.

Finally, the landscape of possible CFs extends beyond time-series. In clinical settings, it also includes text, images and categorical data. These other data types have already received attention in the community. Ultimately, it would be interesting to synchronize the CFs across these data types, e.g., show a clogged artery in the heart from an angiogram, together with an EKG time series reflecting a heart attack, and a text clinical record ``Patient was diagnosed with a myocardial infraction".

\paragraph{Acknowledgements}
{We thank Dr. Rod Tarrago, Dr. Spoorthi K C, Dr. Swathishree Mohan for valuable discussions.} 

\bibliography{references}
\newpage
\appendix
\onecolumn
\section*{Appendix}

\section{Interpretation of the CF-VAE Loss Function}
\label{sec:app_loss_interpretation}

In this section, we interpret the CF-VAE loss function from \Cref{eq:cfvae} by building it up one term at a time. While the most technically interesting use cases for (CF)-VAEs involves datasets with data dimension exceeding the latent dimension, for ease of exposition and visual clarity we use a simple, synthetic $2$-dimensional dataset and a latent space also of dimension $2$.

\begin{figure}[h]
    \centering
    \includegraphics[width=0.95\textwidth]{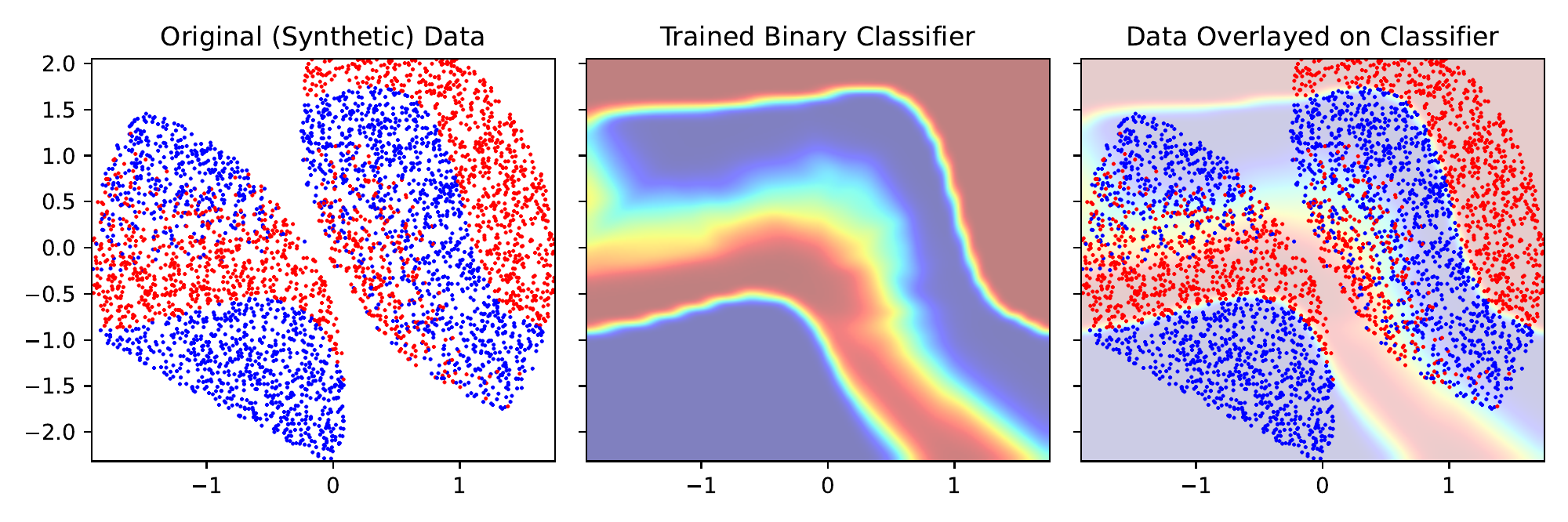}
    \caption{\small{(a) The generated synthetic data. Point colors indicate their labels. Note that this data is largely, but not entirely, separated. (b) The classification probabilities output by a binary classifier trained on this labeled data. (c) The synthetic data overlayed on top of the learned classifier.}}
    \label{fig:deconstruction_synthetic_data}
    \vspace{-.1in}
\end{figure}

In~\Cref{fig:deconstruction_synthetic_data}, we show both the generated synthetic training data and a classifier trained on this data. This classifier will act as the blackbox algorithm used by our CF-VAE. To help with interpretation, the data is constructed to be mostly (but not entirely) separable with a non-trivial shape. The learned binary classifier is a simple feed-forward neural network with 4 hidden layers of 64 neurons each and an un-regularized crossentropy objective.

\begin{figure}[h]
    \centering
    \includegraphics[width=.95\textwidth]{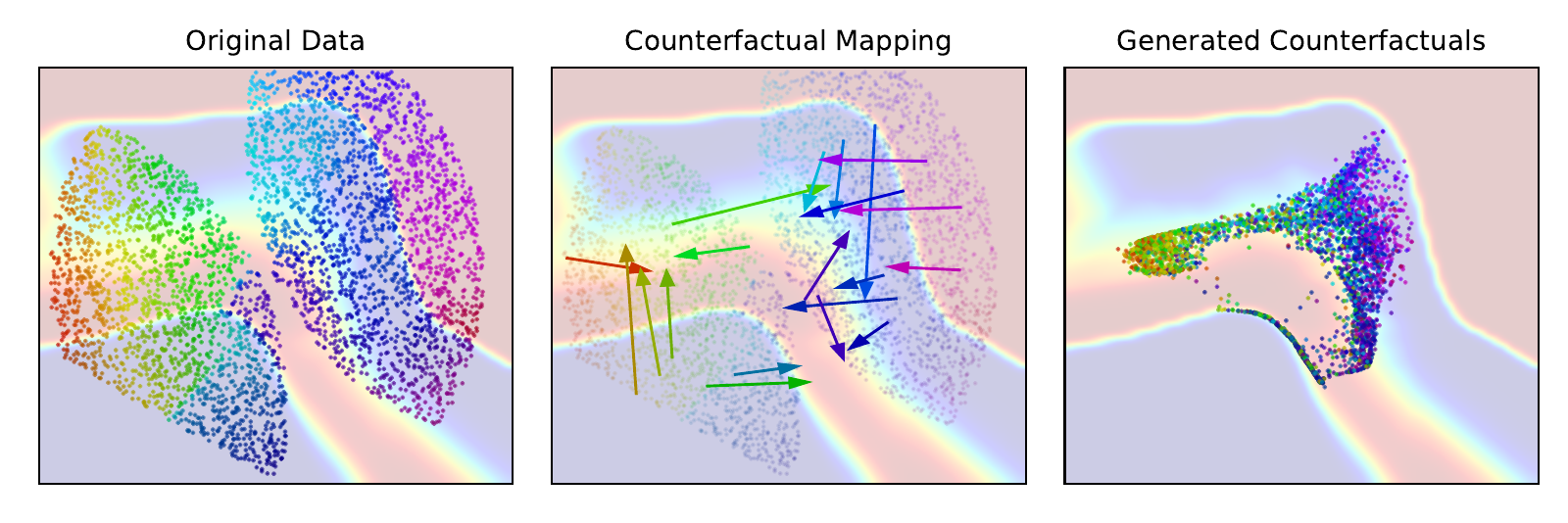}
    \caption{\small{The output of a fully-trained CF-VAE instance with all coefficients equal to $1$. (a) The original synthetic data, recolored and overlayed on top of the binary classifier. (b) A collection of 20 randomly-chosen arrows pointing from the original data point to its generated counterfactual. (c) The locations of all  generated counterfactuals.}}
    \label{fig:deconstruction_all}
\end{figure}

In \Cref{fig:deconstruction_all}, we show the mapping learned by a CF-VAE with all coefficients set to $1$. The left-most plot is similar to \Cref{fig:deconstruction_synthetic_data}(c) in that it shows the 4000 original data points overlayed on the classification boundary, but the label-based coloring is now replaced with a simple gradient. We give points distinct colors so that we can uniquely identify them on the other two plots. In the second plot, we show the counterfactuals chosen by the CF-VAE for 20 randomly chosen sample points: each arrow points from an original point $x$ to its counterfactual $x_{\textrm{cf}}$. Finally, the third plot shows the location of all 4000 counterfactuals, colored to match the original data points. For example, this shows that the violet points on the right of \Cref{fig:deconstruction_synthetic_data}(a) were overwhelmingly mapped to counterfactuals near the center-right of the space. \\

\begin{figure}[h]
    \centering
    \includegraphics[width=.95\textwidth]{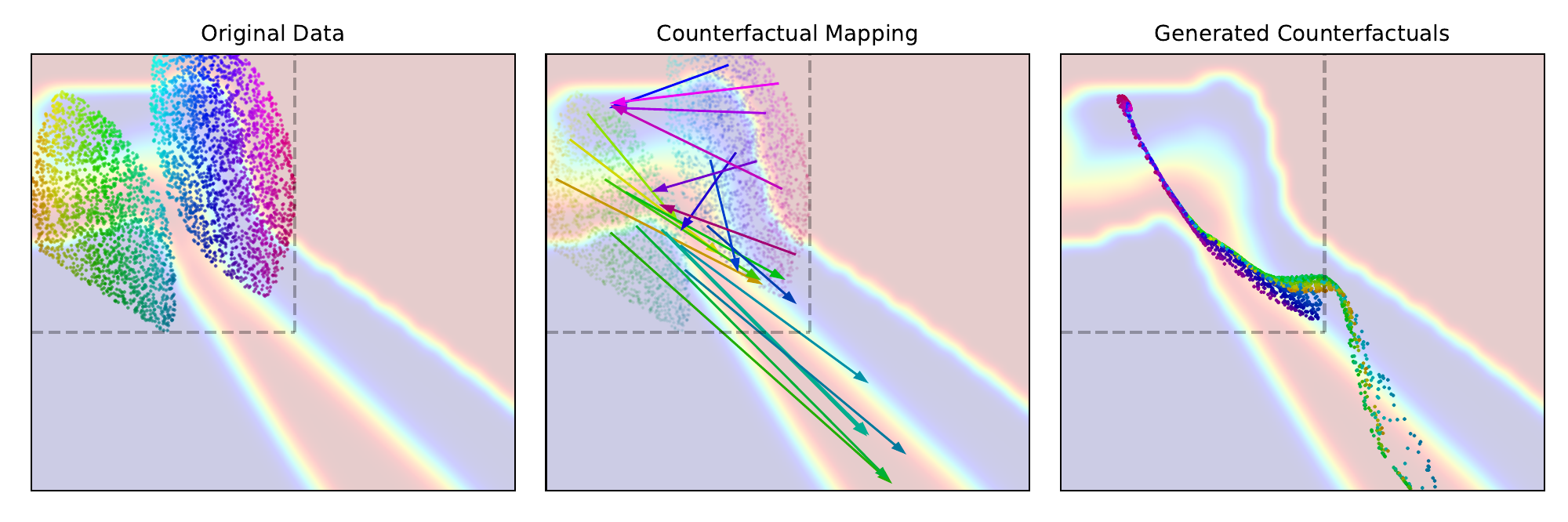}
    \caption{\small{$\textrm{loss} = \textrm{CrossEntropy}(y_{cf}^{prob}, 1-y)$. The output of a network trained only with the crossentropy term. The figure has to be rescaled to show the location of the counterfactual mappings The hyphenated box shows the bounding box of the original data, and we can see many counterfactuals are generated outside of this region.}}
    \label{fig:deconstruction_cf_only}
    \vspace{-.1in}
\end{figure}

\newif\ifforward
\forwardtrue
We now show how the various terms in the loss function build up to this point. As a starting point, we will begin in \Cref{fig:deconstruction_cf_only} with the naive alternative in which the loss function only includes the counterfactual (crossentropy) term from \Cref{eq:cfvae}. In the limit, this is equivalent to setting $\lambda_{cf}$ to an increasingly large number (at which point the CF term dwarfs all other term in the loss).

$$\textrm{loss} = \textrm{CrossEntropy}(y_{cf}^{prob}, 1-y)$$

While most arrows correctly point from red regions to blue regions (or the reverse), we see that these arrows are far longer than those seen in \Cref{fig:deconstruction_all} and, worse, map to counterfactuals far outside of the original data manifold. This is not surprising, as there is no term explicitly constraining the trained network to generate realistic data (outside of the certainty claimed by the trained blackbox model). \\

\begin{figure}[h]
\vspace{-0.15in}
    \centering
    \includegraphics[width=.95\textwidth]{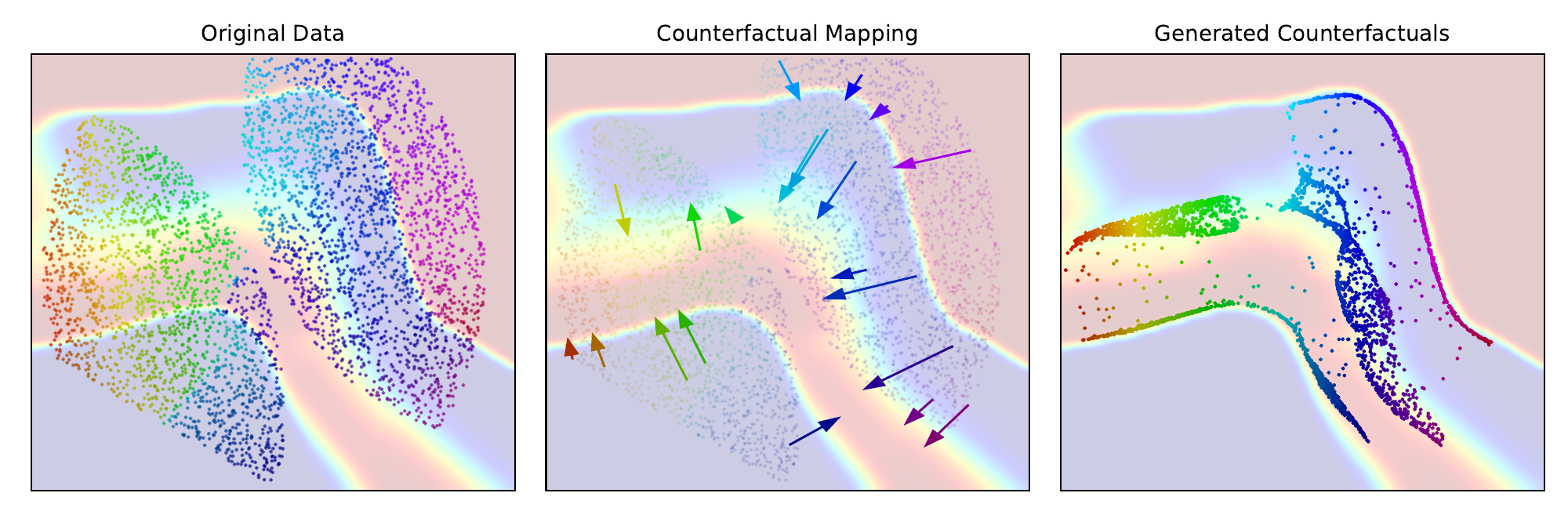}
    \caption{\small{$\textrm{loss} = \textrm{CrossEntropy}(y_{cf}^{prob}, 1-y) {\,+\, ||X -  X_{\textrm{CF}}||_2^2}$. The output of a ``CF-AE'' with only the crossentropy and reconstruction terms.}}
    \label{fig:deconstruction_cf_reconstruction}
    \vspace{-.1in}
\end{figure}

 In \Cref{fig:deconstruction_cf_reconstruction}, we examine what happens if we correct these problems by adding the reconstruction term $||X - X_{\textrm{CF}}||_2^2$ to this loss function.
 
$$\textrm{loss} = \textrm{CrossEntropy}(y_{cf}^{prob}, 1-y) {\color{red}\,+\, ||X -  X_{\textrm{CF}}||_2^2}$$

This shows a drastic improvement over \Cref{fig:deconstruction_cf_only}, but it still suffers from a few flaws. Firstly, we see that the counterfactual manifold is contained almost entirely on the decision boundary of the blackbox classifier. This makes sense and in many cases is useful, as the autoencoder is finding nearby points with a different classification, but borderline patients do not make for striking counterfactuals and the generated CFs suffer from a lack of diversity. Second, we see that the arrows are oftentimes not axis-aligned, meaning that they change multiple parameters at once, which as discussed in \Cref{sec:cfvae} poses a barrier for interpretability.

We address the first concern by adding the regularization term to the loss, thus turning the counterfactual autoencoder into a counterfactual \emph{variational} autoencoder.

$$\textrm{loss} = \textrm{CrossEntropy}(y_{cf}^{prob}, 1-y) \,+\, ||X -  X_{\textrm{CF}}||_2^2{\color{red}\,+\,\mathcal{KL}(\mathcal{N}(\mathbf{\mu_X}, \mathbf{\Sigma_X})|\mathcal{N}(\mathbf{0},\mathbf{I}))}$$

\begin{figure}[h]
    \centering
    \includegraphics[width=.95\textwidth]{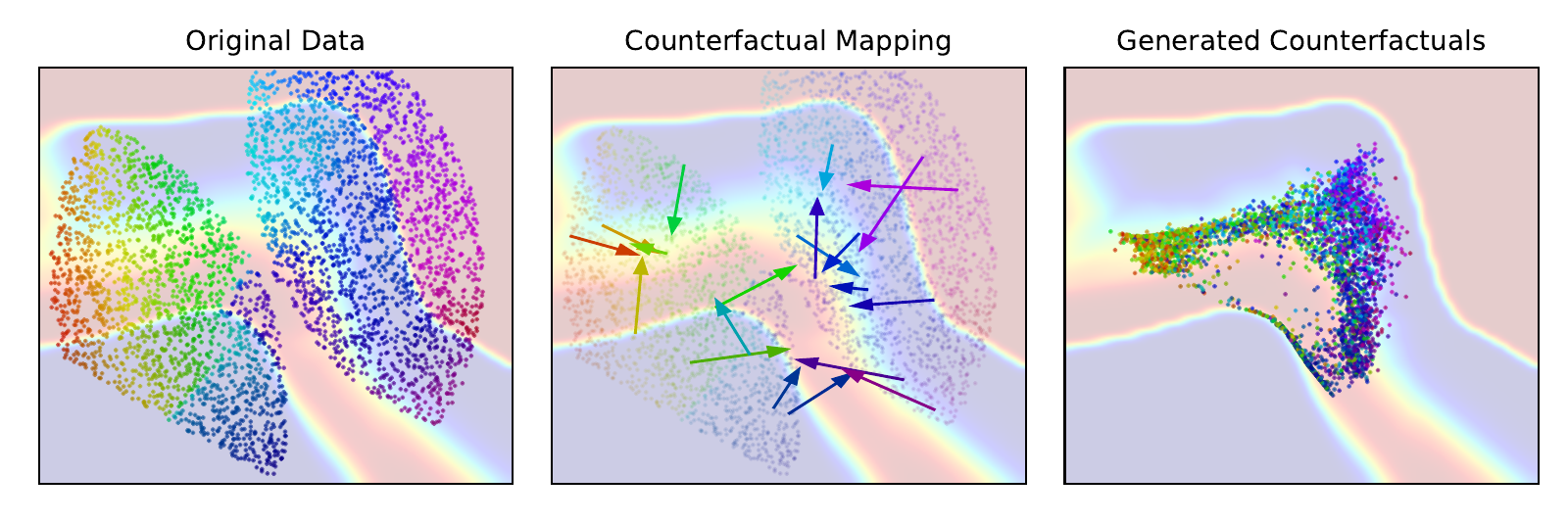}
    \caption{\small{$\textrm{loss} = \textrm{CrossEntropy}(y_{cf}^{prob}, 1-y) \,+\, ||X -  X_{\textrm{CF}}||_2^2{\,+\,\mathcal{KL}(\mathcal{N}(\mathbf{\mu_X}, \mathbf{\Sigma_X})|\mathcal{N}(\mathbf{0},\mathbf{I}))}$. The output of the CF-VAE minus the sparsity term. In contrast to \Cref{fig:deconstruction_cf_reconstruction}, we see more diversity and some motion away from the decision boundary.}}
    \label{fig:deconstruction_cf_reconstruction_kl}
    \vspace{-.1in}
\end{figure}

\Cref{fig:deconstruction_cf_reconstruction_kl} shows the impact of adding the KL term to the objective. This removes the near-zero-width regions from the data, thus allowing for the generation of more diverse counterfactuals that are not all right on the decision boundary.  However, many of the counterfactual directions are still far from axis-aligned.  This is fixed by adding the final sparsity term to the loss function.

$$\textrm{loss} = \textrm{CrossEntropy}(y_{cf}^{prob}, 1-y) \,+\, ||X -  X_{\textrm{CF}}||_2^2\,+\,\mathcal{KL}(\mathcal{N}(\mathbf{\mu_X}, \mathbf{\Sigma_X})|\mathcal{N}(\mathbf{0},\mathbf{I})){\color{red}\,+\,||\mathbf{X} - \mathbf{X_{cf}}||_1}$$

Going back to \Cref{fig:deconstruction_all}, we see that the sparsity term lines the arrows up better with the axes while still maintaining relatively short (and thus meaningful) displacements in counterfactual generation, as well as diversity of counterfactuals. In practice, one would adjust the weights of the various coefficients in the CF-VAE to emphasize (or de-emphasize) the aspects most important to their particular use case. 
\forwardfalse

\forwardfalse

\section{Ranking results}
\label{sec:app_hh}

{The problem of explaining the decision of an ML model to physicians was motivated by the problem of home hospital~\citep{levine2021hospital}. Home hospital is a program in which medical capabilities that would usually be provided in a hospital are brought to the patient's home. It is attractive to both patients and healthcare providers, and is likely to be more relevant as the demand for hospital beds continues to grow~\citep{knight2021provision}. \textit{Selecting the right patients to be sent for home care is critical for this program} - e.g., a patient who might need an intervention in the ICU within the next few hours must not be sent to home care. The current process of assigning a patient to home hospital relies on manual workflows of physicians reviewing data, which is not scalable. ML algorithms can be used to learn effective representations from large datasets and select patients to be sent to home care, \textit{however they need to be {explainable} in order to gain the physician's trust and be deployed in hospitals}. In this paper, we base our ML task on the home hospital problem and frame it using the MIMIC-III dataset~\citep{johnson2016mimic}. We present our CF-VAE, a method to generate counterfactuals for a given binary classifier. In our experiments, we present results of CF-VAE and other prior methods on generating CFs for a model trained on the home hospital task. Note that we choose the publicly available MIMIC-III for our experiments since it facilitates replication. In reality, ICU patients are not candidates for home hospital care. However, the methods proposed in this paper are transferable to a home hospital dataset.}

We frame the home hospital task as a ranking problem where we rank patients based on the predicted time to the next acute intervention. A patient who is likely to require an acute intervention far into the future is more suitable for home hospital when compared to one who might require an acute intervention within the next few hours. We frame this as a pairwise ranking task: given patients $A$ and $B$, we rank them based on who will \textit{first} require a critical intervention. We model the \textit{pairwise} ranking function using a neural network, as proposed in~\cite{burges2005learning}. 

We train a multitask model (Fig.~\ref{fig:pipeline} (a)) to perform two tasks: (1) produce a ranking score, and (2) predict if an acute intervention is required in the next 24 hours. We believe that adding the second task would not hurt the ranking performance (which is the main focus of home hospital) since the two tasks are related. The architecture and training methodology is shown in Fig.~\ref{fig:pipeline} (b). 

The ranking and intervention prediction performance of our model demonstrates the effectiveness of our solution to the home hospital problem. We are able to reliably rank patients, predict interventions, and \textit{generate high quality CFs} to explain the ML model's decision. In this section, we discuss the limitations of our model and some of the opportunities to improve upon it.

In our experiments, we focused on a single intervention such as a vasopressor as an example of an acute intervention. This helped us understand whether vasopressor-related CFs were meaningful.  However, ideally, we would predict whether any acute intervention is needed.  
One complexity that arises is that future acute interventions are influenced by earlier sub-acute interventions. If a condition is caught early enough, then future acute interventions may not be necessary. Hence, a home hospital algorithm when implemented should account for the complex relationship across all interventions. 

\subsection{Ranking and intervention prediction results}
\label{sec:app_additional_experiments}

A pair of 48 hour patient windows A and B are input to the model, and the model produces a ranking order based on who requires acute intervention $I$ first. 
For the purpose of our experiments, we show results on two acute interventions $I$ separately: ventilator and vasopressor.
  
The pairwise ranking and acute intervention prediction performance are shown in Table~\ref{tab:full_rank_results}. We see that the model achieves over 90\% accuracy for both tasks. We also notice that using the entire temporal sequence in the 48 hour window improves the performance of the ranking and prediction task as compared to the slope-intercept representation, indicating that the hourly pattern of the temporal data helps us rank and predict acute intervention more accurately. \textit{For some interventions, such as ventilators, the presence of past interventions is very predictive of similar interventions being required in the near future.} To test this, we perform an ablation study excluding information about the history of the target intervention. We find that while there is a reduction in ranking and prediction accuracy, the model still learns a good representation from only the vitals and other interventions.  

{\begin{table*}[h]
\small
    \centering
    \begin{tabular}{l l l c c c c }
    \toprule
    Experiment & & &  \multicolumn{2}{c}{\textbf{Slope-intercept input}} & \multicolumn{2}{c}{\textbf{Temporal input}}\\
     \hline
     & &  & Vaso & Vent & Vaso & Vent  \\
        \hline 
 \multirow{4}{*}{All features \& } & \multirow{2}{*}{Ranking}  & ACC & $0.88\pm0.01$ & $0.93\pm0.01$ & $0.96\pm0.00$ & $0.94\pm0.00$\\
    
 &  & AUC & $0.95\pm0.01$ & $0.98\pm0.00$ & $0.99\pm0.00$ & $0.98\pm0.00$\\
    
  & \multirow{2}{*}{Intv pred}  & ACC & $0.83\pm0.01$ & $0.83\pm0.12$ & $0.90\pm0.01$ & $0.91\pm0.01$\\
    
 {all patients}&  & AUC & $0.88\pm0.02$ & $0.91\pm0.01$ & $0.96\pm0.00$ & $0.92\pm0.01$\\
 
 \hline
 {Ablation study:} & \multirow{2}{*}{Ranking} &  ACC & $0.89\pm0.00$ & $0.86\pm0.00$ & $0.96\pm0.00$ & $0.88\pm0.00$ \\
        
 {w/o history of }&  &  AUC & $0.96\pm0.00$ & $0.93\pm0.00$&  $0.99\pm0.01$ & $0.94\pm0.01$ \\
 
 {target intv }& \multirow{2}{*}{Intv pred} &  ACC & $0.84\pm0.01$ & $0.79\pm0.01$ & $0.91\pm0.00$ & $0.83\pm0.01$ \\
 
 {as feature}& &  AUC & $0.89\pm0.00$ & $0.83\pm0.00$ & $0.96\pm0.00$ & $0.86\pm0.00$ \\
 \hline 
  {Only patients} & \multirow{2}{*}{Ranking} &  ACC & $0.72\pm0.08$  & $0.95\pm0.00$ & $0.94\pm0.01$ & $0.97\pm0.00$  \\
 {who need }&  &  AUC & $0.80\pm0.00$ & $0.99\pm0.00$ & $0.97\pm0.00$ & $0.99\pm0.00$\\
 {intervention} &  &   &  &  & & \\
  \bottomrule
    \end{tabular}
    \caption{{\small{Multitask model results from three experiments: 1. Ranking and acute intervention prediction using history of vitals, interventions and demographics, 2. Results \textit{without using} the history of the target intervention as an input feature (to rule out any data leakage), 3. Results when we perform ranking only on the patients who require acute intervention within the next 24 hours. \textit{Error bars generated by running the experiment with 10 random seeds.} Here, ACC = Accuracy, AUC = Area under ROC curve, Intv = intervention.}}}
    \label{tab:full_rank_results}
\end{table*}}

\subsection{Additional figures}
\label{sec:app_additional_figures}

\begin{figure}
    \centering
    \includegraphics[scale=0.3]{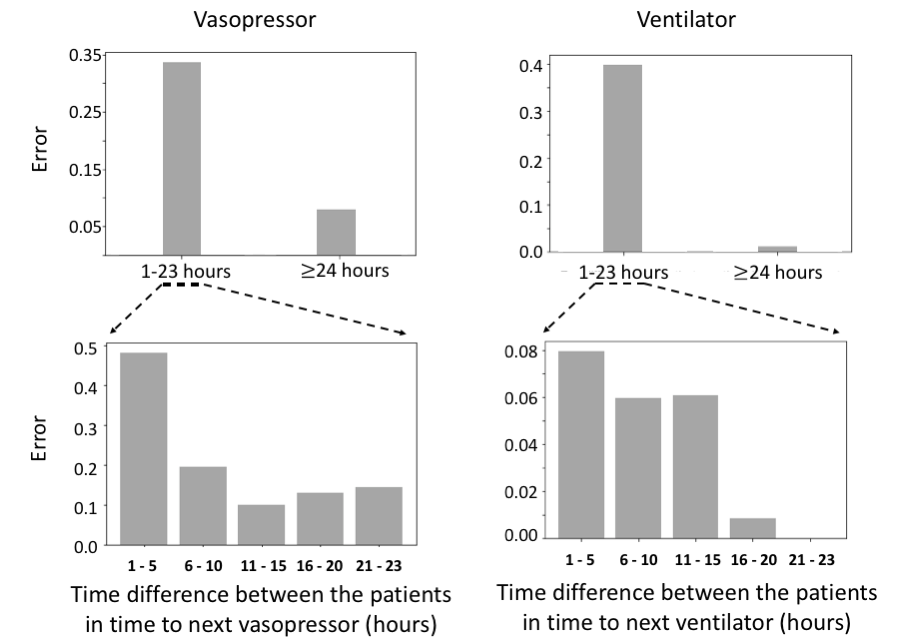}
    \caption{Pair-wise ranking error rate based on the difference in time to intervention between the pairs. In the top figure, the error rate is lower when one of the patients requires an intervention in $>$ 24 hours. This is important since we want to identify such candidates for home hospital.
    The bottom shows ranking performance when evaluated only on pairs where an acute intervention was required within 24 hours - we see a larger error rate when the two patients are $<$ 5 hours apart.}
    \label{fig:rank_error}
\end{figure}

We analyze the errors on the pairwise ranking task and observe that the majority of such mistakes are made when the time difference between the pairs is \textit{small} which is a harder problem (e.g., ranking patients who might need intervention 5hr vs 7hr into the future). Figure~\ref{fig:rank_error} quantifies the error. For the home hospital ranking problem, the most important case is being able to distinguish between patients who require acute-intervention $>24$ hours apart - and our error is low in these cases.

\section{Dataset details}
\label{sec:app_data}

We use the vitals, interventions, and other events recorded from patients in the publicly available MIMIC III dataset \citep{johnson2016mimic} in our experiments. We segment the temporal patient data into 48 hour windows as datapoints to perform ranking and acute intervention prediction over the next 24 hours. We perform a patient-wise split of 70\%-15\%-15\% for training, validation, and testing. We use features corresponding to \textbf{vital signs} ({heart rate, systolic blood pressure, diastolic blood pressure, oxygen saturation, respiratory rate, temperature}), \textbf{interventions} ({ventilator, vasopressor, adenosine, dobutamine, dopamine, epinephrine, isuprel, milrinone, norepinephrine, phenylephrine, vasopressin, colloid bolus, crystalloid bolus}), and \textbf{demographics} ({age, gender}). We use the data pre-processing pipeline in~\citep{wang2020mimic} to transform the MIMIC III raw vital signs and interventions into hourly time series. 

We use the publicly available MIMIC III data set for the experiments in our paper. MIMIC III consists of deidentified data from 53,000 patients admitted to the Beth Israel Deaconess Medical Center in Boston. The temporal patient data is segmented into 48 hour windows, where each window is a data point. For each 48 hour patient window, we have an associated time to next acute intervention ($t_{intv}$) and a binary label of whether they receive an acute intervention within the next 24 hours ($intv_{24}$).  Given the 48 hour window of features for patient A and B, the multitask model produces 3 outputs: 1) pairwise ranking of A, B; 2) prediction that A receives an acute intervention within 24 hours; 3) prediction that B receives an acute intervention within 24 hours. 

To construct the pairs for ranking, we randomly choose two data points that have different values of $t_{intv}$. We randomly choose 50000 pairs of data points for training and 20000 for validation and test. In this paper, we present results on two acute interventions: vasopressor and ventilator. Out of all the patients, 36\% receive a vasopressor and 52\% receive a ventilator. The motivation for our choice of 24 hours as the intervention prediction window is the home hospital design where patients sent to home hospital are visited by a physician at home every day. Hence, we need to ensure that the patient does not crash within 24 hours, before a physician visit.

{
\section{Effect of the weight terms in the loss function}
\label{sec:app_lambda}

The different terms in the loss function are in tension with each other. The VAE loss is optimizing for the reconstructions to lie on the data manifold, whereas the counterfactual loss is trying to produce a reconstruction that lies on the opposite side of the boundary. It is interesting to see how the weight on the counterfactual loss term ($\lambda_{cf}$) affects the CFs in terms of the evaluation metrics considered here. 

In these experiments, we set the sparsity term ($\lambda_S$) to zero, since the effect of the sparsity term  is already demonstrated in Figure 5. We vary the $\lambda_{cf}$ to be between 1 and $10^5$ and evaluate the generated counterfactuals in two ways: 1) Validity: how often the output is of the opposite class, 2) Likelihood: log likelihood of the counterfactual under a KDE fit to the training data.  We show these results in Figure~\ref{fig:kde_acc_lambdacf}(a) and Figure~\ref{fig:kde_acc_lambdacf}(b). 

We observe a competing effect between these methods of evaluation.  On the one hand, as the weight on $\lambda_{cf}$ grows, the more likely the generated point is of the opposite class.  On the other hand, as the weight on the VAE regularization and reconstruction terms grows, the more likely the generated point arises from the KDE on the input space. Hence, we are making a trade-off while selecting the weight for this loss term.

\begin{figure}
    \centering
    \includegraphics[scale=0.5]{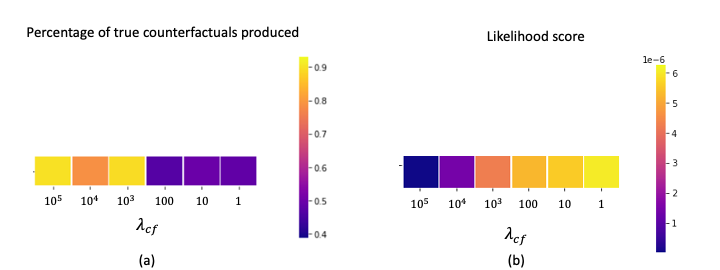}
    \caption{(a) Validity - percentage of produced reconstructions that lie on the opposite side of the boundary when using different values of $\lambda_{cf}$, (b) Log likelihood under a KDE model fit to the training data computed when using different values of $\lambda_{cf}$.}
    \label{fig:kde_acc_lambdacf}
\end{figure}}

\section{Model implementation details}
\label{sec:app_modeldetails}

\begin{figure}[h]
    \centering
    \includegraphics[scale=0.45]{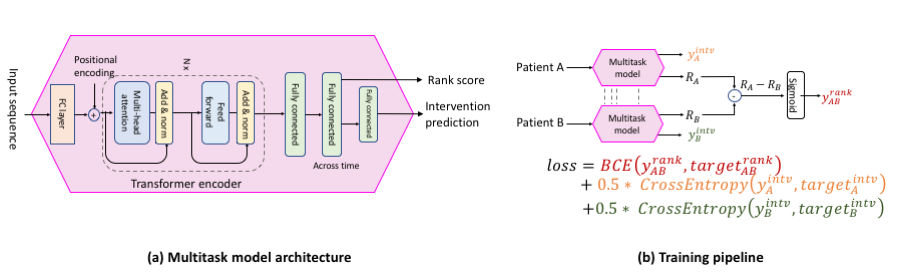}
    \caption{\small{(a) Model architecture for producing ranking and acute intervention prediction output. The architecture consists of a transformer encoder for modeling the temporal sequence of patient data.
    (b) Training pipeline for the ranking and predicting intervention.  }}
    \label{fig:pipeline}
    \vspace{-.1in}
\end{figure}

The data were pre-processed using the pipeline presented in MIMIC-Extract~\citep{wang2020mimic}. 

In the case of the slope-intercept representation, the multitask model was an MLP with 3 layers with ReLU activation and the CF-VAE encoder was an MLP with 4 layers with ReLU activation. 

While using the time series representation, the multitask model consists of a linear embedding layer, a multihead transformer encoder, followed by 3 Fully connected layers. The CF-VAE architecture consists of a linear embedding layer and a multihead transformer layer followed by 1 Fully connected layer.  

The hyperparameters for the multitask model and CF-VAE are presented in Tables~\ref{tab:hyperparam}~\ref{tab:hyperparam_ts}, for both the slope-intercept and time series representation of the vital signs. 
The hyperparameters were chosen by performing a grid search on the validation data. The experiments were performed on an NVIDIA Tesla V100 GPU.

\begin{table}
\centering
\begin{tabular}{ cc }   
(a) Multitask model & (b) CF VAE \\  

\begin{tabular}{ |c|c| } 
\hline
\textbf{Hyperparameter} & \textbf{Value}  \\
\hline
Size of FC layer 1 & 30 \\
\hline
Size of FC layer 2 & 10 \\
\hline
Size of FC layer 3 & 10 \\
\hline
batch size & 32 \\
\hline
learning rate & $1e{-05}$ \\
\hline
epochs & 50\\
\hline
\end{tabular} &  
\begin{tabular}{ |c|c|} 
\hline
\textbf{Hyperparameter} & \textbf{Value}  \\
\hline
Size of FC layer 1 & 40 \\
\hline
Size of FC layer 2 & 100 \\
\hline
Size of FC layer 3 & 60 \\
\hline
Size of FC layer 4 & 30 \\
\hline
batch size & 32 \\
\hline
learning rate & $1e{-03}$\\
\hline
epochs & 50\\
\hline
\end{tabular} \\
\end{tabular}
\caption{Hyperparameters in the slope-intercept representation setting}
\label{tab:hyperparam}
\end{table}

\begin{table}
\centering
\begin{tabular}{ cc }   
(a) Multitask model & (b) CF VAE \\  

\begin{tabular}{ |c|c| } 
\hline
\textbf{Hyperparameter} & \textbf{Value}  \\
\hline
Size of linear embedding layer & 30 \\
\hline
\# transformer encoder layers & 4 \\
\hline
\# heads in multihead self-attention & 5 \\
\hline
Size of FC layer 1 & 10 \\
\hline
Size of FC layer 2 & 20 \\
\hline
batch size & 64 \\
\hline
learning rate & $1e{-04}$\\
\hline
epochs & 100\\
\hline
\end{tabular} &  
\begin{tabular}{ |c|c|} 
\hline
\textbf{Hyperparameter} & \textbf{Value}  \\
\hline
Size of linear embedding layer & 100 \\
\hline
\# transformer encoder layers & 2 \\
\hline
\# heads in multihead self-attention & 2 \\
\hline
Size of FC layer & 100 \\
\hline
batch size & 64 \\
\hline
learning rate & $1e{-05}$ \\
\hline
epochs & 100 \\
\hline
\end{tabular} \\
\end{tabular}
\caption{Hyperparameters in the time series representation setting}
\label{tab:hyperparam_ts}
\end{table}

\begin{figure}[h]
    \centering
    \includegraphics[scale=0.4]{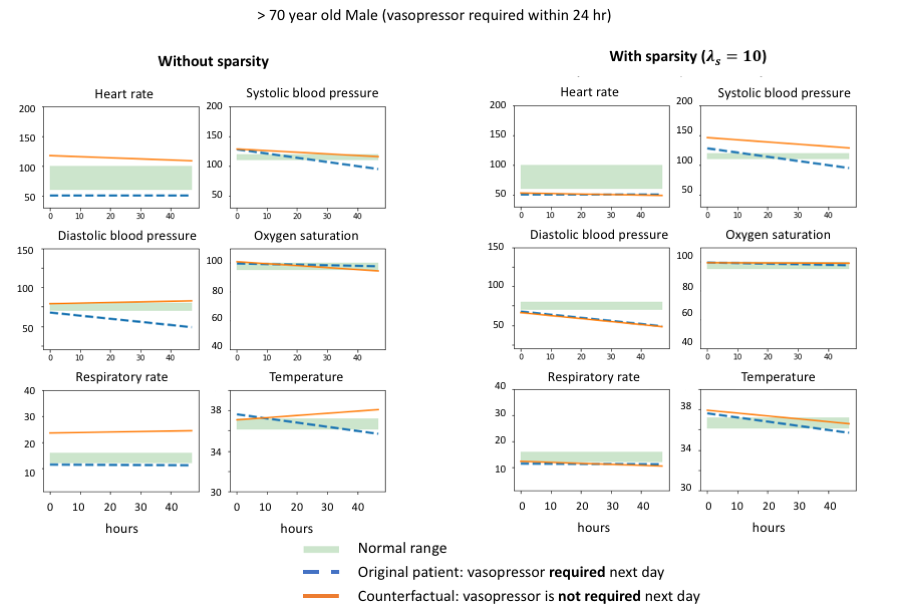}
    \caption{Example 2: Counterfactual produced with and without the sparsity term.}
    \label{fig:sparsity2}
\end{figure}

\section{Proximity of counterfactuals to the original input}
\label{sec:app_proximity}

We can measure the mean squared error between the original input and the generated counterfactual as a measure of the deviation of the CF, which we call proximity. Proximity ensures that we do not produce a very \textit{different synthetic patient} as a counterfactual for a given target patient. Prior methods like DICE directly optimize for proximity at test time. 
CF-VAE does this implicitly via a soft constraint: the log likelihood term Eq.\ref{eq:vae}, under a Gaussian output distribution, reduces to the MSE between the counterfactual and input Eq.~\ref{eq:cfvae}. The method in Delaney, et. al.~\citep{delaney2020instance} encourages proximity by changing only certain regions of the time series and keeping the rest as in the original signal. 
We compute the mean squared error between the original input and generated counterfactual across the test patients for the different methods presented in Table \ref{tab:proximity}. We see that our method outperforms both DICE and Delaney, et. al.

\begin{table}[h]
    \centering
    \begin{tabular}{c c c}
    \toprule
        {} & Ventilator & Vasopressor  \\
        \hline
        {DICE} & 34.48 & 1.8\\
        Delaney et. al. \citep{delaney2020instance} & 4.8 & 0.48 \\
        CF-VAE & \textbf{0.008} & \textbf{0.10}\\
        \bottomrule
    \end{tabular}
    \caption{Proximity (MSE) between CF and input; \textit{\textbf{lower is better.}}}
    \label{tab:proximity}
\end{table}

\section{Comparison to \citep{delaney2020instance} - time series counterfactual}
\label{sec:app_delaney}

\begin{figure}[h]
    \centering
    \includegraphics[scale=0.35]{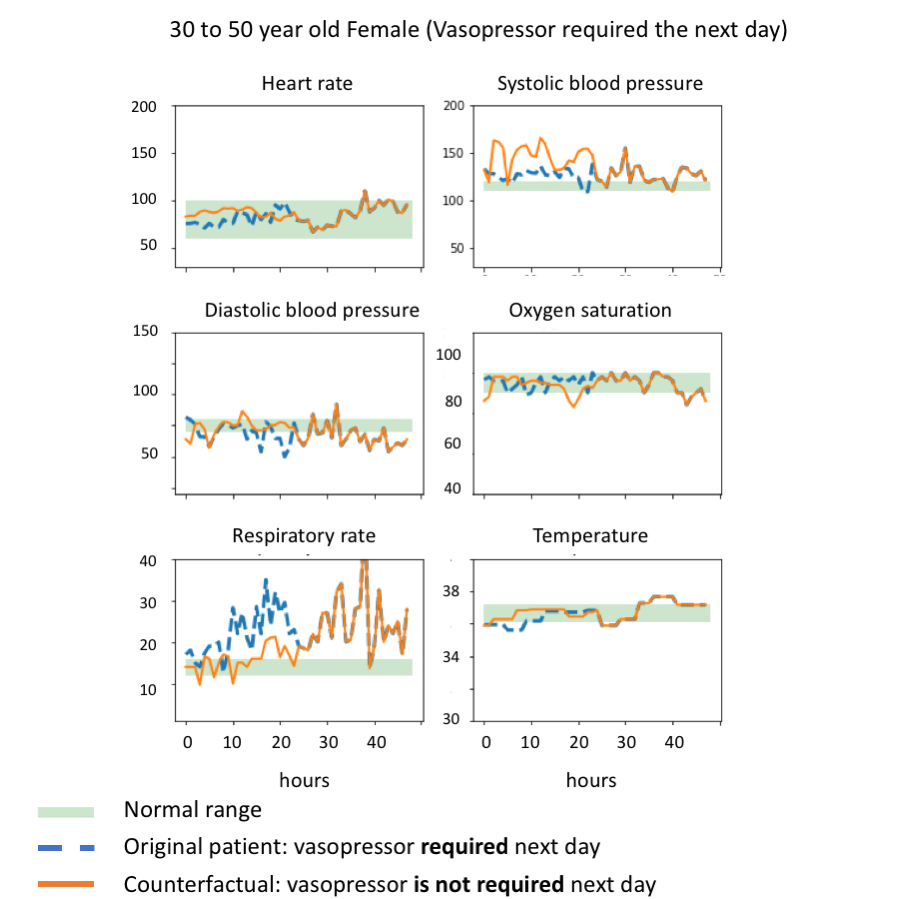}
    \caption{CF generated using the method of Delaney et. al. }
    \label{fig:delaney}
\end{figure}

In \citep{delaney2020instance}, the authors present a method to produce CFs. We believe that this could work for ECG signals (which have been studied in \citep{delaney2020instance}) - replacing a segment (e.g, the QRS complex) of the ECG signal of one patient with the QRS complex of another patient. However, given the amount of variability that exists in vital signs data, combining patients' vital signs can lead to implausible counterfactuals. Figure~\ref{fig:delaney} is an example of a CF generated using \citep{delaney2020instance}.  

We compare the validity and run time of \citep{delaney2020instance} with our time series CF VAE method. We see that \citep{delaney2020instance} has a 100\% validity since the method is designed to at least return the nearest unlike neighbor in the training set as a counterfactual. The training time of CF VAE with time series is higher compared to \citep{delaney2020instance}, which does not require training. However, CF VAE is ~80$\times$ faster than \citep{delaney2020instance} while generating a CF at test time.   

\begin{table*}[h]
    \centering
    \begin{small}
    \begin{tabular}{l c c c c c c}
    \toprule
     & {Ventilator} & {Vasopressor} & &  \\
    \hline
         \small{Method}  & \small{\% CF}   & \small{\% CF}  & \small{Time (s)} & \small{Time (s)} \\
          & \small{validity}  &  \small{validity} & \small{(train)} & \small{(test)} \\
         \hline
         Delaney et. al. \citep{delaney2020instance} & 100\%  & 100\%  & -  & 1.66  \\ 
         CF VAE time series & 95\% & 96\% & 550 & 0.02 \\
        \bottomrule
    \end{tabular}
    \end{small}
    \caption{{Comparison of the validity and run time of Delaney, et. al. \citep{delaney2020instance}} and time-series CF VAE. The validity of \citep{delaney2020instance} is 100\% because at minimum, it returns the nearest unlike neighbor in the training set as a counterfactual. }
    \label{tab:delaney_comparison}
\end{table*}

\section{Visualization of mistakes made by CF VAE}
\label{sec:mistakes}

\begin{figure}
    \centering
    \includegraphics[scale=0.35]{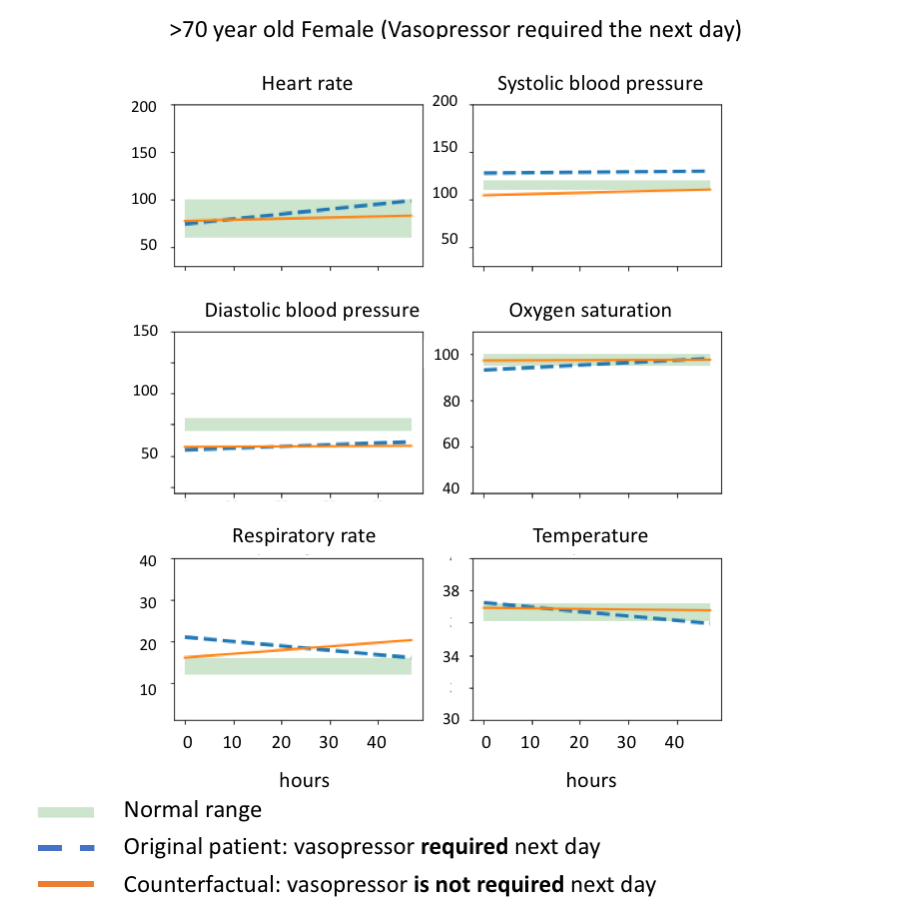}
    \caption{Example of an incorrect counterfactual: the CF indicates that the systolic blood pressure would have to be lowered if vasopressor is not required. Vasopressors are given to increase a patient's blood pressure.}
    \label{fig:mistake}
\end{figure}

Note that the crossentropy loss is minimized to obtain a counterfactual - this does not guarantee a counterfactual 100\% of the time. Additionally, the black-box binary prediction model is also not guaranteed to produce correct labels. See Figure \ref{fig:mistake} for one such example of an incorrect counterfactual. The original patient required a vasopressor the next day. The counterfactual \textit{decreased} the systolic blood pressure for the scenario when a vasopressor would not be required - which does not align with how vasopressors are used clinically. 
We would have to increase the binary prediction model accuracy and increase CF validity of CF VAE to minimize the number of such mistakes.

\newpage
\section{Experiments on images of handwritten digits (MNIST)}
\label{sec:mnist_results}

So far in this paper, we consider clinical time series from MIMIC III and present results based on CFs generated on this data. However, CF VAE is a general solution that can be applied to a given binary classifier to generate counterfactuals. We demonstrate this with experiments on a popular image dataset of handwritten digits (MNIST dataset~\cite{deng2012mnist}). We consider the task of classifying digits 8 vs 3 from handwritten digit images. 

\begin{figure*}
    \centering
    \includegraphics[scale=0.3]{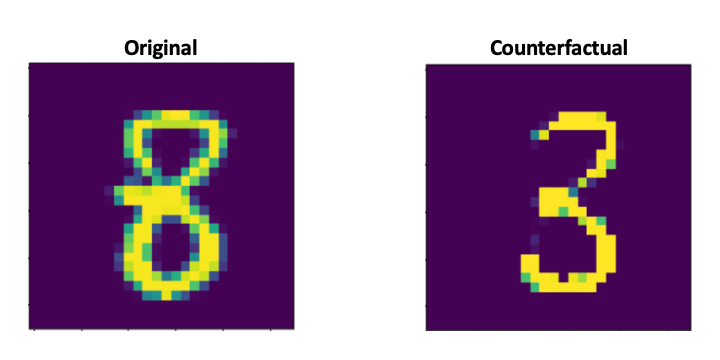}
    \caption{A binary classification model trained to classify images that are 8 vs 3 classifies the original image (shown on the left) as an `8'. The corresponding counterfactual generated is shown on the right. }    
    \label{fig:mnist_eg1}
\end{figure*}

\begin{figure*}
    \centering
    \includegraphics[scale=0.3]{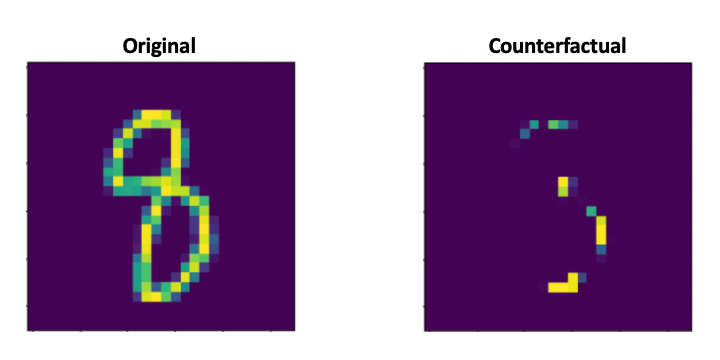}
    \caption{A binary classification model trained to classify images that are 8 vs 3 classifies the original image (shown on the left) as an `8'. The corresponding counterfactual generated is shown on the right. Notice that the counterfactual maintains the style of the original input. }
    \label{fig:mnist_eg2}
\end{figure*}

We train a neural network with linear layers on this task and achieve an accuracy of 99\% for performing this classification given an input image containing the digit.  We then train a CF VAE model to produce a counterfactual image given an original image. The goal here is - given an original image of `8', produce an image that would be classified as `3' under the blackbox classifier. Two example counterfactual images produced are shown in Figure~\ref{fig:mnist_eg1}, \ref{fig:mnist_eg2}. 
Note that, in both cases, CF-VAE creates an image that is no longer an '8' and of a different class.  Moreover, the type and angle of the '3' counterfactual matches the type and angle of the '8'.

\end{document}